# Adversarial Agents on Topology Optimization: Understanding the Fragility and Robustness of Deep Learning-based and Physics-Based Design Models under Adversarial Perturbation

Hoang Anh Nguyen[1], Yuan Hong[2], Hongyi Xu[1*]
[1]School of Mechanical, Aerospace, and Manufacturing Engineering, University of Connecticut, Storrs, CT 06269
[2]School of Computing, University of Connecticut, Storrs, CT 06269
* Corresponding Author: hongyi.3.xu@uconn.edu

**Abstract**

Topology optimization, using both physic-based approaches and deep learning surrogates, serves as a cornerstone for generative design agents in cyber-manufacturing systems. While deep learning surrogates have gained widespread adoption due to their speed in online design generation, this work demonstrates their vulnerability under input perturbations. In this work, we present a mechanics-grounded reliability evaluation framework that formulates an adversarial agent targeting the generative design models. We investigate a strictly non-intrusive threat model where bounded perturbations are introduced exclusively to the initial-density channel, while physical boundary conditions, compliance-gradient channels, network architectures, and solver routines remain intact. Evaluating surrogate models across U-Net, convolutional, and generative architectures with varying physics-gradient conditioning depths demonstrates that bounded initialization noise can cause catastrophic mechanical failure, increasing compliance by multiple orders of magnitude through severed load paths and disconnected supports. Furthermore, we discover that incorporating richer physics-gradient conditioning in the deep learning surrogates does not guarantee monotonic robustness across surrogate families. Finally, physics-in-the-loop recovery demonstrates that initializing the classical SIMP optimizer with perturbed topologies mitigates design performance degradation, having a high probability of restoring compliance to near-baseline levels across tested instances. These findings demonstrate that learned surrogates should serve as physics-verified initializers instead of replacing physics-based solvers entirely in a resilient cyber-manufacturing system. Moreover, the proposed adversarial agent provides a foundation for future training generative design agents robust against noise and targeted perturbations.

---

## 1. Introduction

Topology optimization (TO) has become a cornerstone of modern computational mechanics and automated cyber-manufacturing, enabling the algorithmic synthesis of lightweight, high-performance structural layouts [1-4]. In conventional density-based compliance minimization, numerical optimizers iteratively solve finite-element equilibrium equations and compute adjoint sensitivities to redistribute material throughout a design domain. Although mathematically rigorous, this iterative cycle imposes severe computational bottlenecks when scaling to fine discretizations, multi-load conditions, or real-time generative design pipelines. To alleviate this computational burden, deep learning (DL) surrogates have emerged as autonomous *design agents* capable of near-instantaneous layout prediction [5-10]. These surrogate architectures span encoder-decoder segmentation networks [5, 11], conditional generative adversarial models [8, 12], and physics-informed or latent-space autoencoders [10, 13-16]. Recent advances in machine learning for computational mechanics have further explored physics-based learning strategies [9, 14], implicit neural representations [15], generative graph neural networks for connectivity-guaranteed porous metamaterials [17], uncertainty-aware robust metamaterial optimization [18], manufacturability-aware deep generative design [19], and generative autoencoders with iterative model updating [20], alongside recent CMAME contributions on machine-learning-aided robust topology optimization under uncertainty [21, 22]. By learning rich spatial mappings from problem boundary conditions to converged material distributions, these data-driven models increasingly serve as upstream design generators, real-time feedback engines, or warm-start initializers for numerical solvers.

In connected, cyber-manufacturing systems, overall system resilience relies heavily on the physical trustworthiness of these embedded design agents against cyber-physical anomalies, data tampering, and geometry corruption [23-25]. While recent studies in cyber-physical additive manufacturing have developed geometric and process digital twins for real-time operational resilience [24], established defect control frameworks for in-field fabrication [26], investigated minimal-defect production under material integrity attacks [27], and formulated physics-guided deep learning agents for recovering compromised pre-fabrication digital geometries [28]. Ensuring end-to-end design integrity demands that upstream generative models themselves remain robust to input perturbations. In computer science and adversarial machine learning, foundational studies have established that deep neural networks are remarkably susceptible to bounded, gradient-directed input perturbations, such as the fast gradient sign method (FGSM) and projected gradient descent (PGD), which expose pathological vulnerabilities that stochastic noise fails to uncover [29, 30]. Pitting an adversarial perturbation agent against a learned design agent offers a principled framework for algorithmic vulnerability assessment: systematically exposing worst-case sensitivities reveals how neural networks internally process initialization fields, identifies fragile geometric features, and informs the design of robust initialization schemes and anomaly-filtering defenses.

A central methodological consideration when evaluating learned design agents is formulating a well-defined, physically grounded threat model [23, 25]. In this study, we focus on a strictly **non-intrusive adversary** that perturbs only the initial material density field, $x_{\text{init}}$, under an $L_\infty$ norm bound ($\| \delta \|_\infty \leq \epsilon$), leaving every other physical quantity, network architecture, and solver routine untouched. Section 3 formalizes this threat model, contrasts it with an **intrusive** alternative that instead tampers with the solver itself, and gives the reasons - stealth, feasibility, and clean attribution of any resulting degradation to the surrogate's learned behavior - for concentrating the main study on the non-intrusive channel.

Beyond cyber-manufacturing security, this investigation provides fundamental insights from a computational mechanics perspective. Because density-based topology optimization with SIMP material penalization is inherently non-convex, the optimization landscape contains numerous local minima, and perturbing the initialization density field can steer the ensuing optimization trajectory toward convergence at a different local minimum rather than the intended global (reference) minimum. This sensitivity is most acute for feed-forward neural surrogates, which lack the iterative equilibrium refinement that a classical optimizer uses to escape a poor basin of attraction; consequently, even a minute, visually imperceptible perturbation in the input density space can steer the network to eliminate a critical structural truss, disconnect a reaction support, or introduce ungrounded floating material. While standard machine learning metrics, such as pixel accuracy, mean intersection-over-union, or mean squared error, frequently suggest near-perfect agreement, they are deceptively insensitive to topological continuity and mechanical equilibrium [13, 16]. A tiny localized pixel discrepancy that severs a primary load path causes the structural compliance to explode by multiple orders of magnitude. A rigorous, mechanics-grounded compliance evaluation is therefore essential to assess whether data-driven design agents produce physically viable structures [10, 14].

To systematically examine the physical reliability of learned design agents under non-intrusive input perturbations, this study investigates three central research questions:

- How vulnerable are DL surrogate-based TO models to bounded adversarial perturbations of the initial density channel $x_{\text{init}}$, and does conditioning the network on intermediate compliance-gradient physics enhance its mechanical resilience?
- Does an adversarial initialization crafted against one surrogate architecture also degrade a different, independently trained surrogate under the same attacked input, and if so, is any particular architecture or conditioning depth disproportionately exposed as a transfer target?
- When an adversarial initialization degrades a DL surrogate-based TO model, does rerunning the classical SIMP optimizer from that perturbed state recover structural performance, or does the adversarial perturbation exhibit transferability to the physics-based TO process?

The remainder of this article is structured around addressing these questions. Sections 2 and 3 outline the theoretical formulations of compliance topology optimization, DL surrogate-based TO architectures, and the non-intrusive adversarial attack methodology. Section 4 details the dataset generation procedure and surrogate training protocols. Section 5 presents the empirical investigations: characterizing clean baseline fidelity, analyzing direct surrogate vulnerability and gradient-conditioning effects, assessing cross-model transferability, and evaluating SIMP physics-in-the-loop robustness. Concluding remarks and practical deployment recommendations for resilient cyber-manufacturing are provided in Section 6, with extended sensitivity studies and intrusive threat-model comparisons compiled in the Appendix.

## 2. Technical Background

This section brings together three background fields that form the present study. Their inclusion is deliberate and relational. Density-based topology optimization and the SIMP method provide reference designs, compliance sensitivities, and the physical basis for evaluating structural performance. DL surrogate-based topology optimization uses these reference and physics-derived fields to produce rapid, differentiable topology predictions. Adversarial machine learning then probes the sensitivity of this learned mapping to bounded changes in the software-facing initialization channel. The three areas therefore establish, respectively, mechanical validity, computational acceleration, and reliability assessment rather than serving as independent background topics. Figure 1 visualizes how reference physics informs DL surrogate prediction and how adversarial methods interrogate that prediction. The following subsections provide the corresponding formulations.

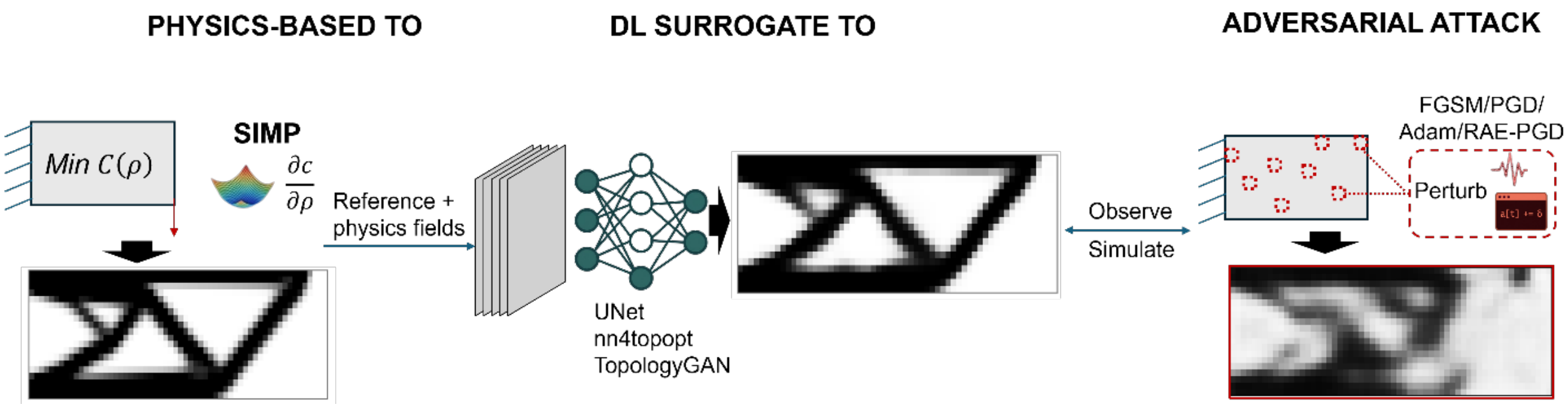

**Fig. 1.** Technical pillars of this work. Physics-based topology optimization supplies the SIMP reference design and compliance sensitivity fields to inform and evaluate the DL surrogate-based design agents (U-Net, nn4topopt, and TopologyGAN). The proposed adversarial agent perturbs the initialization-density channel to test whether bounded input perturbations degrade designs generated by both physics-based and DL surrogate-based models.

### *2.1 Density-based topology optimization*

We consider standard minimum-compliance topology optimization on a discretized design domain with $n$ elements and element densities $\rho = [\rho_1, \dots, \rho_n]^T \in [0, 1]^n$ [1-4]:

$$\min_{\rho \in [0,1]^n} C(\rho; p) = F(p)^T u(\rho; p),$$
$$\text{subject to } K(\rho; p) u(\rho; p) = F(p), \ \frac{1}{n}\sum_{e=1}^{n} \rho_e \le v(p), \tag{1}$$

where $K(\rho; p)$, $u(\rho; p)$, $F(p)$, and $v(p)$ denote the global stiffness matrix, nodal displacements, external loads, and target volume fraction, respectively. Intermediate densities are penalized via the SIMP interpolation [2]

$$E(\rho_e) = E_{\min} + \rho_e^{p_{\text{pen}}}(E_0 - E_{\min}) \tag{2}$$

where $E_0$ and $E_{\min}$ denote the solid and near-void (minimum) Young's moduli, respectively, and the penalization exponent $p_{\text{pen}} = 3$ discourages intermediate, non-physical densities. The adjoint compliance sensitivities, given by

$$\frac{\partial C}{\partial \rho_e} = -p_{\text{pen}} \rho_e^{p_{\text{pen}}-1}(E_0 - E_{\min}) u_e^T k_0 u_e \tag{3}$$

are regularized via standard convolution filtering [3, 4] prior to optimality-criteria updates [31]. While numerical SIMP iteratively resolves finite-element equilibrium to reach reference topology $y_{\text{SIMP}}$, deep learning surrogates predict $\hat{\rho}$ in a single feed-forward pass without intermediate equilibrium enforcement.

*2.2 DL surrogate-based TO architectures*

Learned topology-optimization surrogates turn a repeated optimization problem into a prediction problem. A trained network receives a structured description of the design case and returns a candidate density field without running a new optimization loop. In abstract form, such a surrogate can be written as

$$f_\theta : s(p) \mapsto \hat{\rho} \tag{4}$$

where $s(p)$ may contain supports, loads, volume information, and sometimes physics-derived fields, depending on the model. The predicted field $\hat{\rho} \in [0, 1]^n$ can then be used directly as a candidate topology or passed to a physics optimizer for checking or refinement. This study evaluates three established feed-forward surrogate families - a U-Net encoder-decoder, an nn4topopt-style convolutional network, and a TopologyGAN-style generative model - chosen because they represent distinct architectural paradigms while sharing a common grid-based input-output representation. Fig. 2 provides a simplified schematic of all three architectures and the multi-channel input tensor.

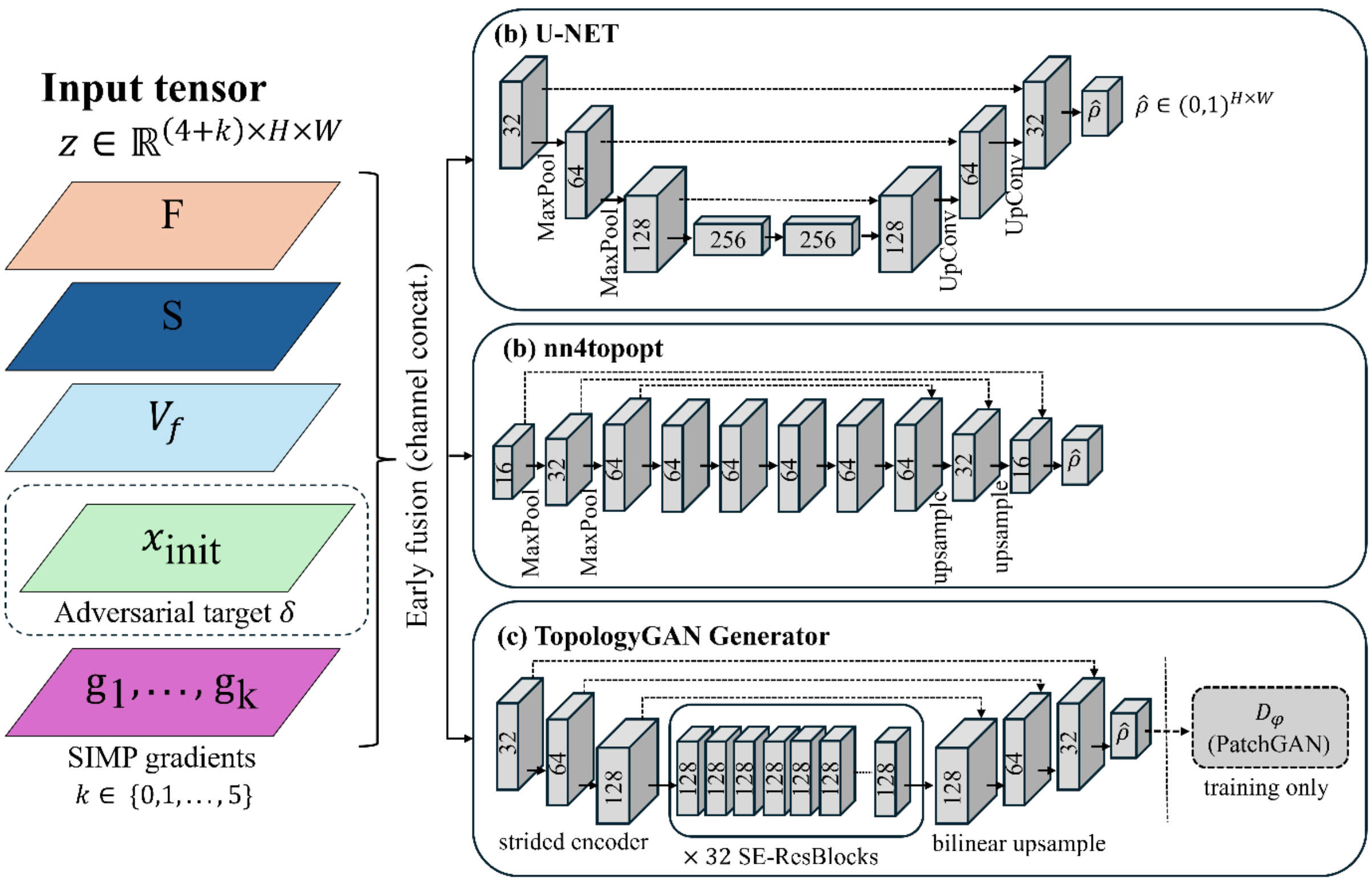

**Fig. 2.** DL surrogate architectures and multi-channel physics-gradient injections. Left: the input tensor $z \in \mathbb{R}^{(4+k)\times H\times W}$ with boundary-condition channels, the initial density $x_{\text{init}}$ (adversarial perturbation target, dashed border), and $k$ compliance-gradient fields from SIMP iterations. (a) U-Net encoder-decoder with skip connections. (b) nn4topopt fully convolutional network. (c) TopologyGAN generator with SE-residual blocks and PatchGAN discriminator (training only). All channels enter via early fusion at the first convolutional layer.

**Multi-channel input and physics-gradient injection**

All three architectures receive the same multi-channel spatial input tensor on an $H \times W$ finite-element grid:

$$z = [\underbrace{F, S, V_f}_{\text{boundary conditions (3 ch.)}}, \underbrace{x_{\text{init}}}_{\text{initial density (1 ch.)}}, \underbrace{g_1, \ldots, g_k}_{\text{gradient channels } (k \text{ ch.})}] \in \mathbb{R}^{(4+k)\times H\times W} \quad (5)$$

where $F$ encodes the applied load (nonzero at the loaded element, zero elsewhere), $S$ marks kinematic fixed-support constraints, $V_f$ broadcasts the target volume fraction as a constant spatial field, and $x_{\text{init}} \in [0, 1]^{H\times W}$ is the initial density field (the adversarial perturbation target). When gradient conditioning is used ($k \geq 1$), the channels $g_1, \ldots, g_k$ contain element-wise compliance sensitivities $\partial C/\partial \rho_e$ from the first $k$ SIMP iterations.

Gradient conditioning follows an **early-fusion** strategy in all three architectures: the sensitivity fields are concatenated along the channel axis at the network input, so the first convolutional layer receives $4 + k$ channels. This means the physics information modulates feature extraction from the very first convolution, without separate encoder branches or intermediate injection points. In the main experiments, $k \in \{0, 1, 5\}$, yielding four, five, or nine input channels, respectively.

**U-Net encoder-decoder (UNet)**

The U-Net [11] follows a four-level encoder-decoder structure with skip connections. Each level uses a double-convolution block ($3 \times 3$ convolutions, batch normalization, ReLU). The contracting encoder downsamples through three $2 \times 2$ max-pooling operations with channel widths $32 \rightarrow 64 \rightarrow 128$, and a bottleneck block operates at $256$ channels. The expanding decoder uses $2 \times 2$ transposed convolutions ($128 \rightarrow 64 \rightarrow 32$), concatenating upsampled features with corresponding encoder features via skip connections to preserve fine geometric detail. A $1 \times 1$ convolution with sigmoid activation produces $\hat{\rho} \in (0, 1)^{H \times W}$.

**Fully convolutional network (nn4topopt / NN)**

The nn4topopt architecture [5] is a lighter three-level encoder-decoder ($16 \rightarrow 32 \rightarrow 64$ channels) with no batch normalization. It uses $3 \times 3$ convolutions with ReLU activations, spatial dropout ($p = 0.1$) at the second encoder and decoder levels, and a four-layer $64$-channel bottleneck. The decoder upsamples via nearest-neighbor interpolation rather than transposed convolutions, concatenating with encoder skip features at each level. With approximately 135K parameters (versus 1.93M for U-Net), the normalization-free design represents a different capacity–regularization trade-off.

**Conditional generative adversarial network (TopologyGAN / GAN)**

TopologyGAN [8] frames topology prediction as conditional image generation. The **generator** is a deep SE-Res-UNet: a three-level strided encoder ($5 \times 5$ convolutions, stride $2$, batch normalization, leaky ReLU; $32 \rightarrow 64 \rightarrow 128$ channels), followed by $32$ sequential squeeze-and-excitation residual blocks at $128$ channels. Each residual block applies two $5 \times 5$ convolutions with batch normalization and a channel-attention SE module (reduction ratio $r = 16$) before the additive skip connection. The decoder uses bilinear upsampling, $5 \times 5$ convolutions, and encoder skip connections, with sigmoid output. The **PatchGAN discriminator** (three strided $5 \times 5$ layers, $32 \rightarrow 64 \rightarrow 128$ channels) classifies spatial patches as real or fake during training only; at inference, only the generator is used. With approximately 6.79M parameters and $5 \times 5$ kernels throughout, TopologyGAN has a substantially larger receptive field than the $3 \times 3$-kernel architectures.

All three models share the same feed-forward deployment: a learned model proposes a topology, after which a physics optimizer may verify or refine it [6, 9, 10]. Because the early-fusion input and feed-forward prediction provide no intermediate physics-based correction, any perturbation to $x_{\text{init}}$ propagates through the entire network unchecked, the vulnerability channel investigated in Section 3.

Recent generative work raises the standard for learned TO evaluation. Diffusion, latent-diffusion, and flow-matching approaches broaden the learned topology-generation landscape and emphasize performance-aware guidance, generalization, and conditioning-signal choice [12, 32-34]. These newer models may follow different conditioning strategies from the architectures above, but they reinforce the same evaluation issue: a learned topology should be judged by the mechanics of the resulting design, not only by its visual similarity to a reference.

### *2.3 Adversarial attack formulations*

These attack algorithms originate in the computer-vision and adversarial machine-learning literature, where they were first used to expose brittle decision boundaries in image classifiers [29, 30]; this paper imports them, largely unmodified, as reproducible ways of constructing a bounded perturbation to one input channel of a mechanics surrogate. This use is distinct from robust topology optimization: projection-based robust TO and recent ML-aided robust TO instead optimize a design against a prescribed uncertainty model at design time [21, 22, 35]. This paper does not solve a robust optimization problem. It keeps a trained surrogate fixed after training and asks whether a bounded, adversarially chosen perturbation to one software-facing input channel can change the mechanical quality of the predicted layout.

The four attack algorithms are chosen to represent common first-order ways of constructing such a bounded input perturbation for a differentiable model, a one-shot signed-gradient step, its iterative counterpart, a free-variable adaptive optimizer, and a noise-averaged variant, each detailed below. The goal is not to catalogue every possible attack; black-box, combinatorial, and solver-intrusive attacks would answer different questions.

Let $\delta$ denote the perturbation to the input channel being attacked and let $\mathcal{L}(\delta)$ be the attack loss defined in Section 3.3. Each method searches within the same box constraint,

$$\parallel \delta \parallel_{\infty} \leq \epsilon \tag{6}$$

that is, every perturbation element is restricted to the interval $[-\epsilon, \epsilon]$. The perturbed density channel is also clipped to the valid range $[0, 1]$.

**Fast Gradient Sign Method (FGSM):** FGSM takes a single signed-gradient step [29] and is the earliest of the four methods to be proposed in the adversarial machine-learning literature. Rather than moving along the raw gradient $\nabla_{\delta}\mathcal{L}$, it moves along its element-wise sign, so every element of $x_{\text{init}}$ is pushed by exactly the same magnitude $\epsilon$ and only the *direction* of the local gradient is used:

$$\delta_{\text{FGSM}} = -\epsilon \, \text{sign}\big(\nabla_{\delta}\mathcal{L}(0)\big) \tag{7}$$

This makes FGSM a single backward pass through the surrogate, essentially free to compute, and gives a useful lower bound on how vulnerable a model is. Its main limitation is that it takes only one linear step from the clean input, so it can underestimate the worst-case perturbation whenever the loss surface curves noticeably within the $\epsilon$-ball, which motivates the iterative methods below.

**Projected Gradient Descent (PGD):** PGD repeats the same signed-gradient idea over multiple smaller steps of size $\alpha \ll \epsilon$, re-evaluating the gradient at the current perturbed point and projecting the result back onto the allowed $\ell_\infty$ ball after every step [30]:

$$\delta^{m+1} = \Pi_{\mathcal{B}_\epsilon}\left(\delta^m - \alpha \operatorname{sign}\left(\nabla_\delta \mathcal{L}(\delta^m)\right)\right) \tag{8}$$

where $\mathcal{B}_\epsilon$ denotes the elementwise interval $[-\epsilon, \epsilon]$. By re-linearizing the loss at each intermediate point rather than only at $\delta = 0$, PGD can follow local curvature across the ten steps used in this study (Section 4) and is widely treated as a much stronger worst-case probe than a single FGSM step, at the cost of ten forward/backward passes instead of one.

**Adam-based Perturbation Optimization:** This variant drops the fixed-magnitude sign step entirely and instead treats $\delta$ as a free variable optimized with adaptive moment estimation [36], tracking a running first and second moment of the gradient for every element of $x_{\text{init}}$ independently, followed by box clipping to $[-\epsilon, \epsilon]$ after each of the 20 update steps used here (Section 4). Because the effective step size is rescaled per element rather than shared uniformly across the whole density field, this variant can make faster progress where the local loss surface is flat and slower, better-controlled progress where it is steep, which the fixed-magnitude sign methods above cannot do.

**Resampled / Expectation-Over-Transformation (RAE-PGD):** RAE-PGD follows the same iterative PGD update but replaces the single-sample gradient $\nabla_\delta \mathcal{L}(\delta^m)$ with an average over several noisy re-evaluations of the attacked channel, three samples per step with noise standard deviation 0.05 in this study (Section 4), before taking the signed step [37]. Averaging over resampled noise gives a smoother, lower-variance estimate of the local descent direction, so a failure found by RAE-PGD is less likely to be an artifact of one particular gradient evaluation at one particular point and more likely to reflect a broader, noise-robust weakness of the surrogate's decision surface around $x_{\text{init}}$.

## 3. Methodology

This section formulates the proposed adversarial threat model against DL surrogate-based TO. It first consolidates why the study focuses on a non-intrusive adversary rather than an intrusive one, then formalizes the adversarial initialization model (Section 3.2), the attack objective (Section 3.3), and the compliance-ratio metrics (Section 3.4) used throughout the results.

### 3.1 Why a non-intrusive threat model

The adversary studied throughout this paper interacts solely through the external, software-facing data-exchange boundary: it perturbs only the initial material density field, $x_{\text{init}}$, under an $L_\infty$ norm bound ($\| \delta \|_\infty \leq \epsilon$), while every physical quantity, boundary-condition encoding, gradient-conditioning channel, network architecture, model weight, and optimizer routine remains exactly as specified by the original design case. We deliberately contrast this setting with an **intrusive** adversary that instead tampers with quantities inside the numerical solver itself, such as the compliance-sensitivity field, the sensitivity filter, or the applied load vector. Appendix A.5 reports a control experiment in which these intrusive channels are attacked directly on the Top88/SIMP optimizer: sensitivity poisoning, phantom load injection, and filter poisoning each drive the optimizer to a severely degraded or non-convergent final compliance (up to three orders of magnitude above the clean baseline), because they corrupt the information the optimizer itself relies on to re-solve equilibrium at every iteration. Intrusive attacks are therefore not a weaker threat than the non-intrusive channel studied here, if anything, they are more damaging once realized.

The paper nonetheless concentrates its main empirical study on the non-intrusive channel for three practical reasons. An $x_{\text{init}}$-only perturbation never touches the solver's governing equations, sensitivity computation, or filtering step, so it evades the kind of solver-level sanity check, equilibrium residual growth, filter-output bounds, load-vector checksums, that a defender might place around the optimization core. Sensitivity or filter tampering cannot make the same claim, since it directly corrupts the quantities those checks are designed to monitor; the non-intrusive channel is, in this sense, inherently more **stealthy**. It is also **easier and cheaper** to reach than the solver's internals: an initialization, proposal, or warm-start density field is exactly the kind of artifact that is commonly cached, exchanged between upstream design tools, or supplied by a separate service in a generative design pipeline. An attacker therefore needs only write access to that upstream data boundary, not to the solver's own, often proprietary or access-controlled, source code, internal sensitivity arrays, or optimizer update rule. Finally, constraining the perturbation exclusively to $x_{\text{init}}$ yields a clean **attribution**: any resulting structural degradation can be traced unambiguously to the surrogate's learned inductive bias rather than to external numerical manipulation of the physics itself. Together, these properties make the non-intrusive channel the more realistic and more diagnostic threat to study when the object of interest is the *learned surrogate*, even though the intrusive channel, once available to an attacker, is already known from Appendix A.5 to be capable of defeating the physics solver as well.

Fig. 3 illustrates the non-intrusive threat model, emphasizing that the attack modifies only $x_{\text{init}}$ while all physical and conditioning channels remain fixed. The structured inputs, attack objective, and paired compliance measures shown in the figure are formalized in Sections 3.2-3.4.

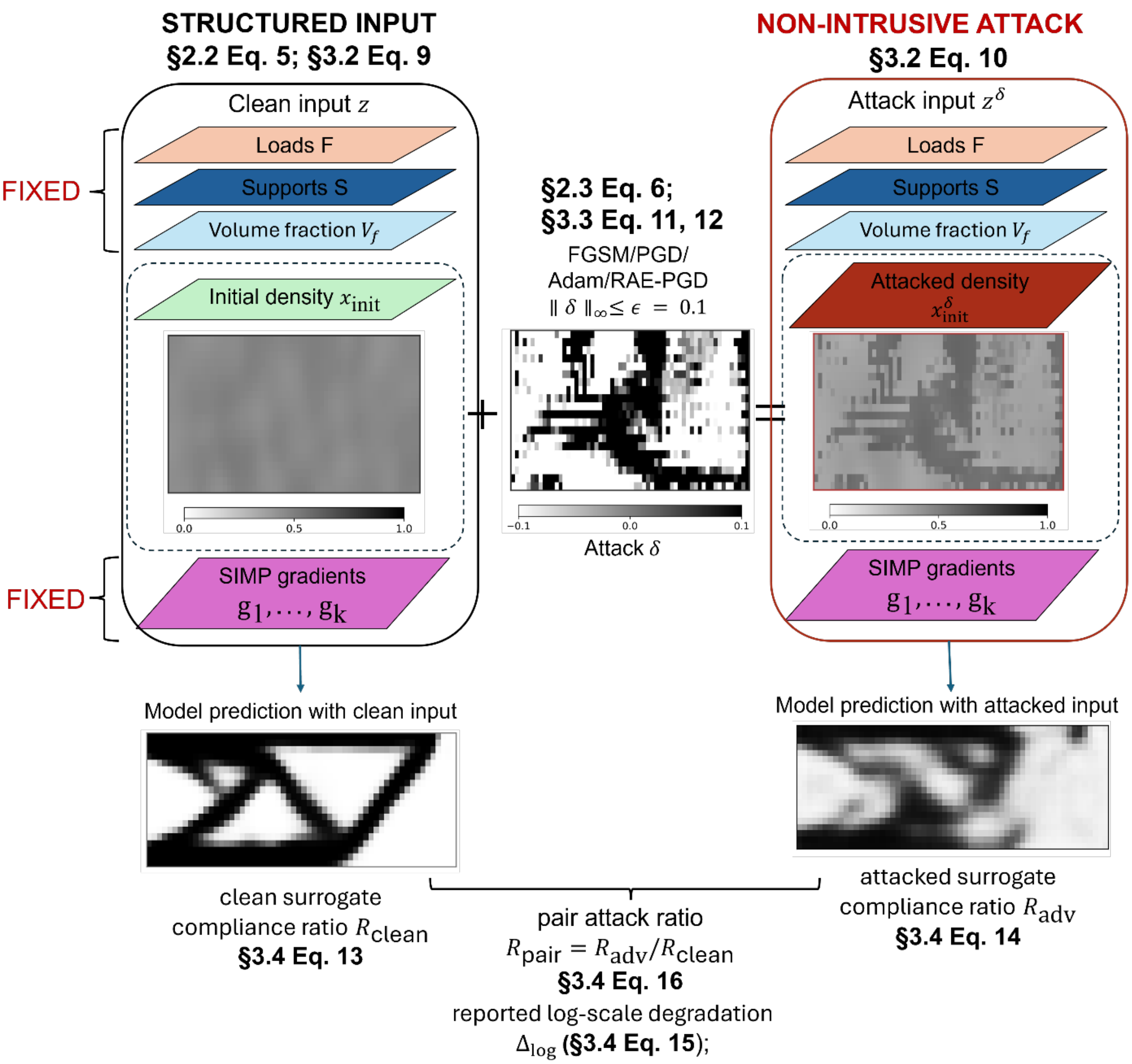

**Fig. 3.** Non-intrusive attack construction and paired evaluation. The clean structured input contains the physical channels, initial density, and SIMP-gradient fields. The attack adds a bounded perturbation $\delta$ only to $x_{init}$ to form $z^\delta$; the load, supports, volume fraction, gradient channels, model weights, and physical problem remain fixed. The source surrogate is evaluated with the clean and attacked inputs to obtain $R_{clean}$ and $R_{adv}$, from which the paired attack ratio $R_{pair}$ and log-scale degradation $\Delta_{log}$ are computed.

*3.2 Adversarial initialization model*

The non-intrusive adversary formalized above perturbs only the added initial-density channel, $x_{init}$, in the augmented surrogate interface used for this evaluation; the original physical problem is not changed. For compact notation, let $x_0$ denote this $x_{init}$ field. Given a clean input

$$z = (b, (p), x_0, g_1, \dots, g_k) \tag{9}$$

the adversarial input is

$$z^{\delta} = \left(b, (p), \Pi_{[0,1]}, (x_0 + \delta), g_1, \dots g_k\right) \tag{10}$$

where $\Pi_{[0,1]}$ clips element densities to the valid density range. The perturbation is bounded by $\| \delta \|_{\infty} \leq \epsilon$. Unless otherwise stated, $\epsilon = 0.10$. This is a controlled 10% density-range perturbation setting, not a universal mechanical tolerance. Appendix A.4 shows how the direct-attack results change at smaller and larger budgets.

One consequence of this scope is worth stating precisely: for models with gradient-conditioning channels, keeping those channels fixed while changing $x_{\text{init}}$ deliberately creates a channel-consistency check. If a deployment recomputes all gradient-like fields after any initialization update, the present results should not be transferred to that workflow without new experiments. In this sense, the model isolates one reliability channel, learned-surrogate sensitivity to bounded changes in the initialization-density input, rather than every way a deployment could vary.

The realism of this channel depends on the deployment setting. The paper assumes a workflow in which an initialization, proposal, or warm-start density field may be supplied, reused, or generated by upstream software before a learned surrogate or physics optimizer is invoked. This assumption is consistent with the dataset and model inputs used here and with physics-refined surrogate workflows that use learned outputs as initial guesses [10]. The assumption defines a scoped reliability probe, not a claim that all ML-assisted topology-optimization deployments expose the same channel.

*3.3 Attack objective and scope*

For each source surrogate, the attack is generated in the surrogate input space by modifying only $x_{\text{init}}$, using one of the four first-order procedures already formalized in Section 2.3 (FGSM, PGD, Adam-based optimization, and RAE-PGD): each searches for a bounded perturbation $\delta$, with $\| \delta \|_{\infty} \leq \epsilon$, that drives the surrogate output toward a chosen adversarial target; the target itself, and how a candidate perturbation is scored once generated, are specified next. Let the clean prediction be $\hat{\rho}_0 = f_{\theta}(z)$. The direct attacks use a heuristic output-space adversarial target

$$t = \Pi_{[0,1]}(1 - \hat{\rho}_0) \tag{11}$$

and optimize the perturbed input so that the surrogate prediction moves toward this target:

$$\mathcal{L}_{\text{attack}}(\delta) = \left\| f_{\theta}\left(z^{\delta}\right) - t \right\|_2^2, \quad \| \delta \|_{\infty} \leq \epsilon \tag{12}$$

Each of the four candidate attack procedures minimizes this same loss under the same perturbation constraint; they differ only in how the descent direction on $\delta$ is computed, the sign-gradient step of FGSM and PGD (Section 2.3, Eqs. 7-8) versus the adaptive-moment and noise-averaged variants used by the Adam-based and RAE-PGD procedures, respectively (also Section 2.3). The complement target is not a compliance-gradient attack and is not claimed to be mechanically optimal; it is a reproducible target used to test whether bounded $x_{\mathrm{init}}$ changes can move a surrogate toward high-compliance predictions.

The attack-generation loss is not the reported mechanics metric. After each candidate perturbation is generated, the attacked surrogate prediction is evaluated by compliance under the original physical problem $p$. For each source model, the selected direct attack is the candidate with the strongest median log-scale compliance-ratio increase ($\Delta_{\log}$, Section 3.4, Eq. 15) on the evaluation set.

*3.4 Evaluation metrics*

Let $C(\cdot; p)$ denote compliance evaluated under the original physical problem. The clean surrogate compliance ratio is

$$R_{\mathrm{clean}} = \frac{C(\hat{\rho}_0; p)}{C(y_{\mathrm{SIMP}}; p)} \tag{13}$$

The attacked surrogate compliance ratio is

$$R_{\mathrm{adv}} = \frac{C(\hat{\rho}^{\delta}; p)}{C(y_{\mathrm{SIMP}}; p)} \tag{14}$$

The reported log-scale degradation is

$$\Delta_{\log} = \log_{10}(R_{\mathrm{adv}}) - \log_{10}(R_{\mathrm{clean}}) \tag{15}$$

The paired attack ratio is

$$R_{\mathrm{pair}} = \frac{R_{\mathrm{adv}}}{R_{\mathrm{clean}}} = \frac{C(\hat{\rho}^{\delta}; p)}{C(\hat{\rho}_0; p)} = 10^{\Delta_{\log}} \tag{16}$$

It compares attacked and clean predictions for the same model and physical problem.

The log-scale metric is needed because some thresholded predicted structures can become disconnected or nearly singular, producing very large raw compliance ratios. These large ratios are treated as failure indicators under the stated numerical convention rather than as precise engineering load ratings. For the same reason, the results report the median of $R_{\mathrm{pair}}$ across the test set, not the mean: a handful of near-singular cases would dominate a mean and make it uninformative, whereas the median summarizes the typical outcome. The exceedance counts reported alongside the median (Section 5.2-5.4, Appendix A.2-A.3), the fraction of test cases exceeding a given ratio, separately capture how often the tail actually occurs.

## 4. Dataset and training of DL surrogate-based TO models

All experiments use a compliance topology-optimization dataset on a $20 \times 60$ finite-element grid. The dataset contains 300 samples. Each sample includes boundary/load/volume-condition channels, the added initial density field $x_{\mathrm{init}}$, up to five compliance-gradient channels ($k \in$ $\{0, 1, 2, 3, 4, 5\}$, as defined in Eq. 5), and a SIMP-optimized target topology. The seed-0 split contains 210 training samples, 45 validation samples, and 45 test samples.

The dataset generator fixes the left boundary degrees of freedom and samples the target volume fraction uniformly from $[0.3, 0.7]$. It applies a unit downward load to a randomly selected admissible node on the lower, upper, or right boundary region. The initial density field is initialized at the target volume fraction, perturbed by Gaussian-smoothed random variation with amplitude 0.05, and clipped to $[0.001, 1.0]$. The dataset metadata records 300 accepted samples from 300 attempts and no solver failures.

The model set is intentionally limited to the three feed-forward architectures introduced in Section 2.2, a U-Net-style encoder-decoder, an nn4topopt-style convolutional model, and a TopologyGAN-style generator. The study is therefore a controlled reliability evaluation of these selected surrogates, not a comprehensive benchmark of every learned topology-optimization model.

The experiments adapt these architectures to the common multi-channel input interface defined in Eq. 5. This augmentation is part of the evaluation design, not a requirement of the original U-Net, nn4topopt, or TopologyGAN papers, and it lets the physical problem stay fixed while one initialization-like channel is varied.

The experimental labels combine architecture type and conditioning depth. “UNet”, “NN”, and “GAN” denote the U-Net, nn4topopt-style, and TopologyGAN-style models introduced in Section 2.2, and the suffix $k$ denotes the number of compliance-gradient channels added to the augmented input. Thus, UNet0, NN0, and GAN0 use no gradient channels; UNet1, NN1, and GAN1 use one compliance-gradient channel ($k = 1$); and UNet5, NN5, and GAN5 use five compliance-gradient channels ($k = 5$). The suffix denotes conditioning depth only, not a distinct architecture variant.

To keep the main comparisons readable, the principal tables and result figures report $k = 0$, $k = 1$, and $k = 5$ for each architecture. These nine models span no gradient conditioning, shallow conditioning, and the deepest tested conditioning.

The U-Net models use trained weights from the fixed-ablation runs with base width 32, MSE loss, Adam at learning rate $10^{-3}$, cosine annealing to $10^{-5}$, batch size 16, and 40 epochs. The nn4topopt-style and TopologyGAN-style models were retrained under one convergence-controlled protocol with a common training seed. Their objective combines binary cross entropy with soft Dice loss at weight 0.5; optimization uses Adam at learning rate $10^{-3}$, batch size 16, and no explicit weight decay. Training permits at most 200 epochs. ReduceLROnPlateau halves the learning rate after 12 stale validation epochs down to $10^{-5}$, and early stopping begins after epoch 60 with patience 40 and minimum improvement $10^{-4}$. The best validation-loss model is retained. Afterward, each source surrogate is evaluated with FGSM, PGD with ten steps, Adam-based perturbation optimization with 20 steps and learning rate 0.03, and RAE-PGD with three noisy gradient-accumulation samples per step and noise standard deviation 0.05.

## 5. Results and findings

This section evaluates robustness in four stages, moving from clean surrogate fidelity to direct vulnerability and then to cross-model transfer and physics-based recovery. This organization distinguishes degradation within the attacked source surrogate from effects that persist when the same initialization is evaluated by another model or re-optimized with SIMP. Figure 4 provides a roadmap for reading the resulting analyses.

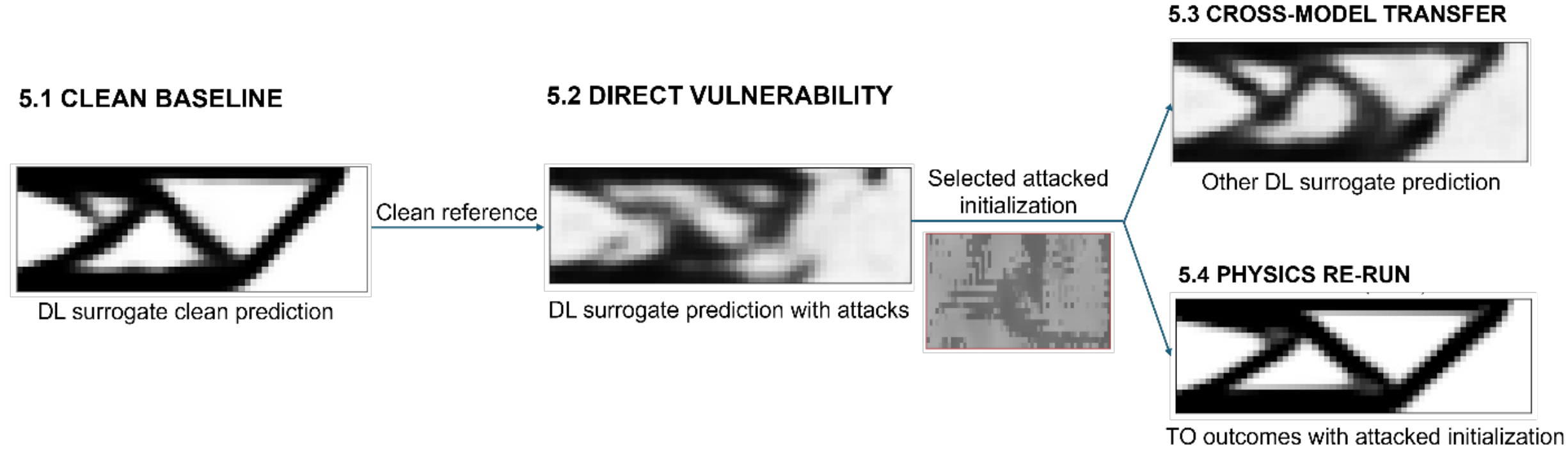

**Fig. 4.** Roadmap of the Section 5 results. The clean surrogate prediction establishes the baseline in Section 5.1, and the direct evaluation in Section 5.2 identifies the selected attack for each source model. The same selected attacked initialization then branches into two complementary tests: prediction by another DL surrogate in the cross-model transfer analysis of Section 5.3 and physics-based topology optimization in the re-run analysis of Section 5.4. Loads, supports, volume targets, and evaluation metrics remain fixed across these comparisons.

### *5.1 Results of Clean DL Surrogate-based Designs*

Across all nine architecture-depth combinations, clean surrogate predictions closely track the SIMP reference in both image-space agreement and mechanical compliance, establishing a stable baseline before any adversarial perturbation is introduced. Fig. 5(a)–(b) traces this image-space and mechanics-space performance as the number of gradient-conditioning channels increases; all depths from zero through five are plotted and labeled on the x-axis to reveal the progression within each architecture, while filled markers identify the depths used in the main comparisons. Fig. 5(c) then compares clean predictions on one representative physical problem, holding the initialization density, force, supports, volume target, and SIMP reference identical across every row so that only the surrogate architecture and gradient-conditioning depth change.

Per-model clean metrics are reported in Appendix Table A.1. Across the nine models used in the main comparisons, clean image agreement is consistently high: binary accuracy is about 0.93 to 0.95, IoU is 0.87 to 0.91, and F1 is 0.93 to 0.95. Clean compliance is also typically close to the SIMP reference, with median clean/GT ratios from $0.999\times$ to $1.047\times$. The nn4topopt-style IoU improves with gradient conditioning, while the U-Net and TopologyGAN-style trends are mildly nonmonotone. In the shared-setup comparison of Fig. 5(c), the right-edge loading produces a full-span design that uses the complete horizontal domain; all nine predictions preserve the principal load path, and their case-specific clean/GT ratios range from $0.98\times$ to $1.05\times$. The panel titles report both absolute compliance, $C$, and the thresholded prediction-to-SIMP ratio in parentheses.

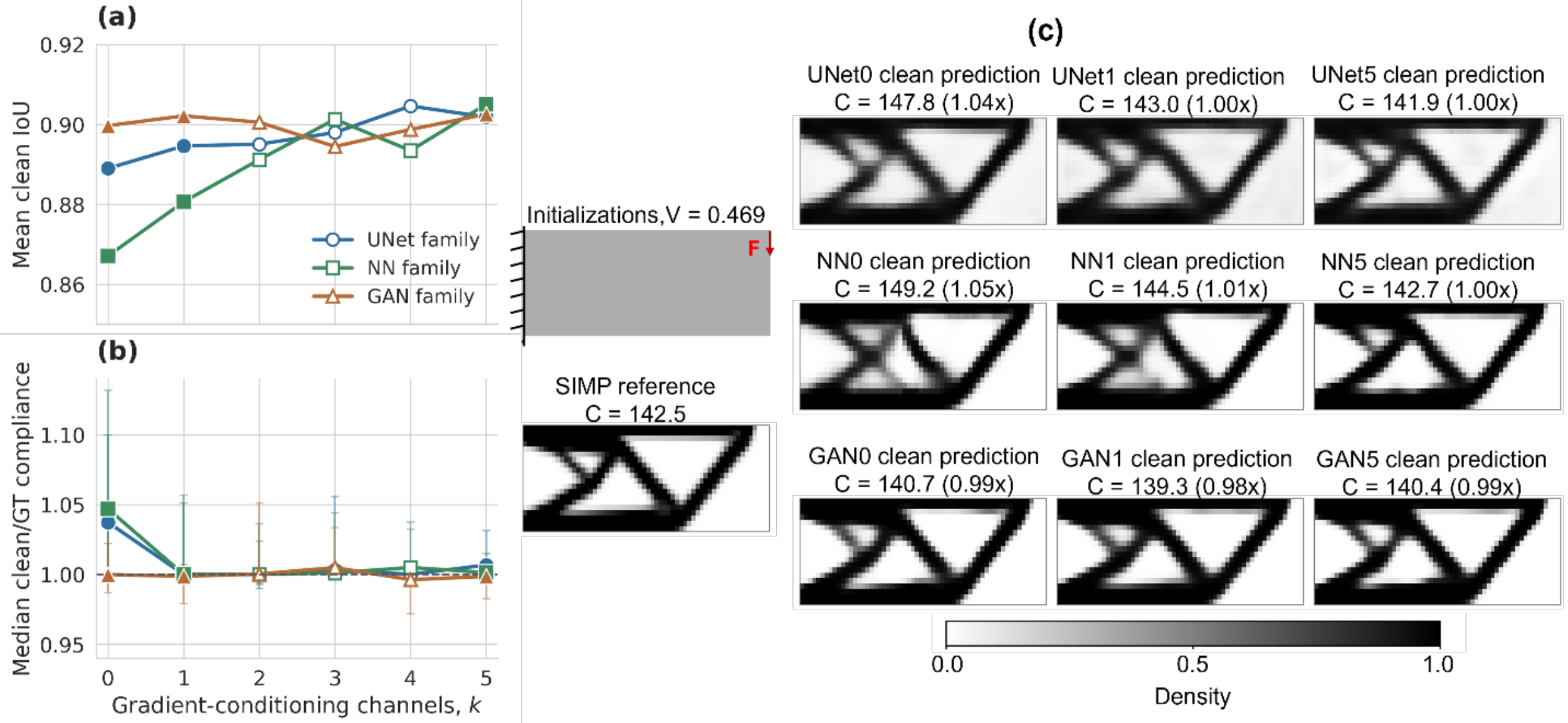

**Fig. 5.** Clean surrogate performance. (a) Mean clean IoU versus gradient-conditioning channels, $k$. (b) Median thresholded clean/GT compliance versus $k$; the dashed line marks agreement with the SIMP reference. (c) Clean predictions for a common load/support setup: rows correspond to surrogate architectures; columns show the shared $x_{\text{init}}$, predictions at $k = 0,1$, and 5, and the SIMP reference.

*5.2 Direct Initialization-Attack Vulnerability*

We first ask whether bounded input perturbations under the $L_\infty$ norm ($\| \delta \|_\infty \leq \epsilon = 0.10$, a 10% change in the density range) applied only to $x_{\text{init}}$ can lead some surrogates to propose very different load paths, even though the load, supports, volume target, and all other inputs are unchanged. Table 1 compares four attack optimizers across the three surrogate architectures and the three depths used in the main comparisons; each entry reports the median paired change in thresholded compliance relative to the same model's clean prediction.

The degradation is concentrated rather than uniform, and it is severe wherever it occurs. The clearest U-Net failure appears at UNet1, where PGD raises the median paired compliance by $132.65\times$; this is not a rare tail event but the typical outcome for that configuration, since Appendix Table A.2 shows that 31 of the 45 held-out test problems exceed a $2\times$ compliance increase, 26 exceed $10\times$, and 25 exceed $100\times$. The TopologyGAN-style models show a similar but less concentrated pattern: GAN1 reaches $36.19\times$ under PGD (35/45 cases exceed $2\times$, 27/45 exceed $10\times$) and $34.15\times$ under Adam, while GAN0 is also sensitive under Adam ($4.63\times$ median; 24/45 exceed $2\times$, 21/45 exceed $10\times$). In contrast, the nn4topopt-style medians stay close to $1.0\times$ for all four attacks, and the corresponding exceedance counts remain low across the family (at most 10/45 test cases above $2\times$ and 1/45 above $10\times$ for any nn4topopt configuration in Appendix Table A.2). The attack channel therefore exposes vulnerable trained models within the tested set rather than a uniform collapse of every surrogate, and where a model is vulnerable, the failure recurs across a majority of the test set rather than appearing only in isolated cases.

Inspecting the attacked layouts in Fig. 6 clarifies the mechanism behind these numbers. The clean layouts carry an unbroken load path from the support to the loaded node, whereas the attacked UNet1 and GAN1 predictions lose the connecting structural member entirely and become nearly disconnected under the thresholded compliance calculation; the resulting free-floating or singular segments carry negligible stiffness, which is why the thresholded compliance ratio explodes by six to seven orders of magnitude in this example ($1.26 \times 10^7\times$ for UNet1, $6.66 \times 10^6\times$ for GAN1) rather than degrading gracefully. The NN1 example changes much less, consistent with the aggregate pattern in Table 1. This behavior is consistent with the non-convex SIMP landscape discussed in Section 1: because feed-forward surrogates enforce no iterative equilibrium constraint at inference time, a bounded perturbation to $x_{\text{init}}$ can push the prediction into a qualitatively different, structurally disjoint local pattern rather than merely perturbing member sizes within the same topology.

The response is also not monotone with the number of gradient-conditioning channels: UNet1 and GAN1 (one gradient channel) are markedly more fragile than UNet0/GAN0 (no gradient channel) or UNet5/GAN5 (five gradient channels) within the same architecture family (Table 1). Because gradient conditioning changes what the surrogate has learned to weight during training rather than adding an equilibrium check at inference time, one additional sensitivity channel appears to create a narrow, more exploitable dependency for the U-Net and TopologyGAN-style architectures, whereas five channels return to more redundant, harder-to-exploit conditioning; the nn4topopt-style family, by contrast, remains comparatively insensitive at every tested depth. The current evidence therefore does not support a claim that adding more gradient channels is a general robustness mechanism; robustness instead appears to depend on an interaction between architecture and conditioning depth that is specific to each model family.

The bold entries in Table 1 identify the attack selected per model, carried forward into the later examples and comparisons. Appendix A.4 reports how the selected attacks change with the perturbation budget.

**Table 1.** Direct initialization attacks for the nine reported models. Entries are the median paired ratio of attacked to clean thresholded compliance, $R_{\text{pair}} = C_{\text{adv}}/C_{\text{clean}}$ (Eq. 16), over the 45 held-out test problems. A ratio above $1.0\times$ means the attacked prediction is mechanically worse (higher compliance, i.e., lower stiffness) than the same model's clean prediction; a ratio at or below $1.0\times$ means the attack produced no measurable degradation. Bold entries mark the attack selected per model (Section 3.3), which is carried forward to Fig. 6 and Tables 2-3. FGSM: fast

gradient sign method (Eq. 7); PGD: projected gradient descent (Eq. 8); Adam: Adam-based perturbation optimization; RAE-PGD: resampled/expectation-over-transformation PGD.

| Family | Attack | $k = 0$ | $k = 1$ | $k = 5$ |
| --- | --- | --- | --- | --- |
| UNet | FGSM | $0.92\times$ | $1.35\times$ | $1.11\times$ |
| | PGD | $1.48\times$ | $132.65\times$ | $1.23\times$ |
| | Adam | $1.20\times$ | $105.84\times$ | $1.37\times$ |
| | RAE-PGD | $1.38\times$ | $39.18\times$ | $1.38\times$ |
| NN | FGSM | $1.00\times$ | $1.09\times$ | $1.05\times$ |
| | PGD | $1.03\times$ | $1.16\times$ | $1.10\times$ |
| | Adam | $1.03\times$ | $1.15\times$ | $1.11\times$ |
| | RAE-PGD | $1.00\times$ | $1.16\times$ | $1.07\times$ |
| GAN | FGSM | $1.10\times$ | $7.10\times$ | $1.03\times$ |
| | PGD | $2.51\times$ | $36.19\times$ | $1.07\times$ |
| | Adam | $4.63\times$ | $34.15\times$ | $1.04\times$ |
| | RAE-PGD | $3.03\times$ | $34.60\times$ | $1.06\times$ |

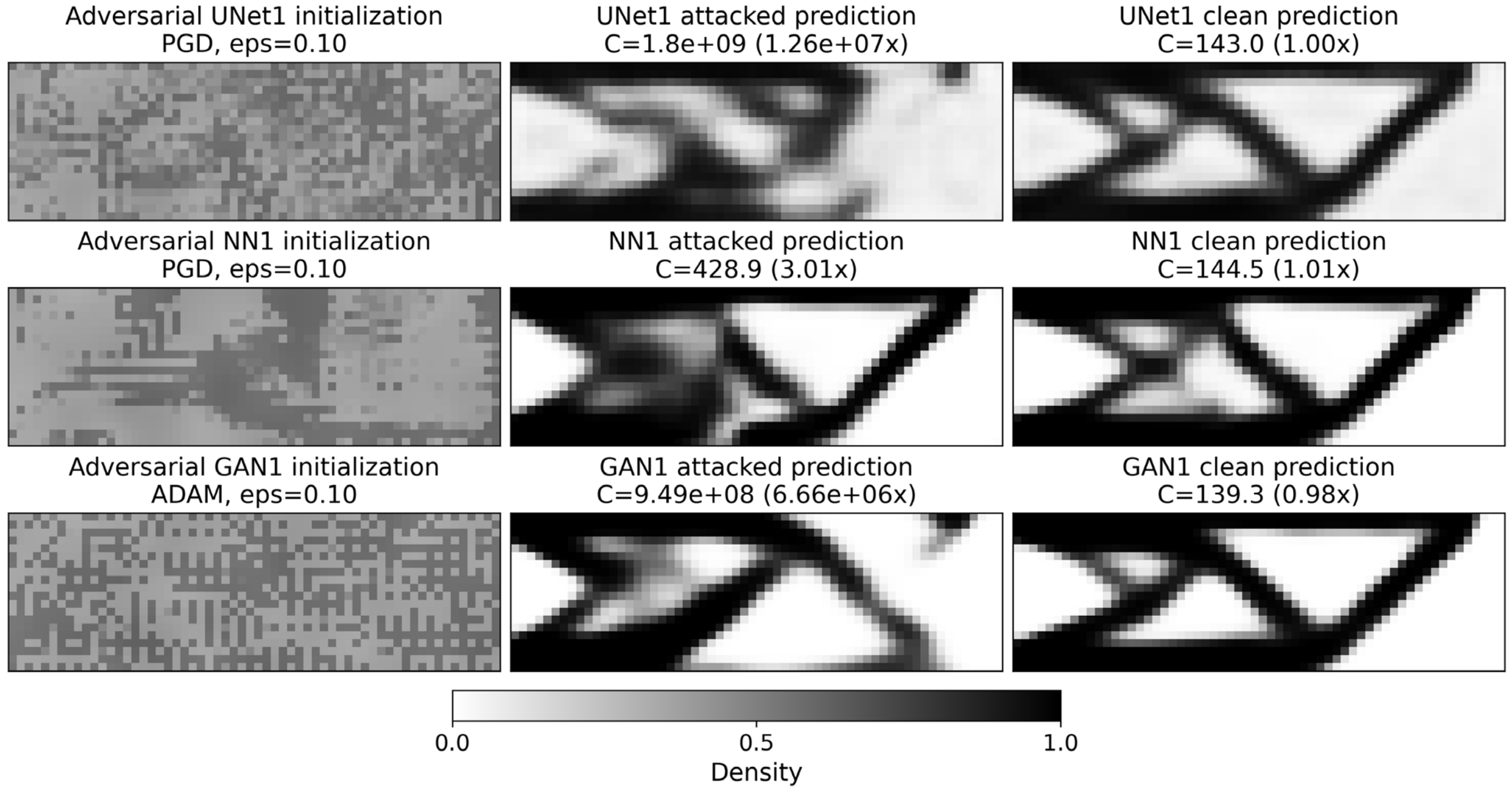

**Fig. 6.** Direct attack examples on the shared setup. Each row shows the adversarial initialization, the attacked prediction, and the clean prediction for the same source model.

*5.3 Transfer Across Surrogate Targets*

In Section 5.2, each model is targeted by a dedicated adversarial agent using the approach introduced in Sections 2.3 and 3.2-3.3. In this section, we examine attack transferability across models to evaluate whether perturbations trained on one surrogate degrade another. Specifically, the same attacked $x_{\text{init}}$ is passed unchanged to other trained surrogates while the load, supports, target volume, and conditioning channels remain fixed. We take a target compliance increase greater than $1.05\times$ (a $5\%$ or larger rise in thresholded compliance) as the threshold for a *notable* transfer effect, since this exceeds the typical clean-model variability reported in Appendix Table A.1.

Table 2 shows that transfer is limited in the median and strongly target-sensitive. The bold diagonal cells reproduce the direct attacks from Table 1. Away from the diagonal, most medians remain close to $1.0\times$ and below the notable-transfer threshold; the largest off-diagonal median is NN1 → UNet1 at $1.16\times$. Thus, the strongest direct failures do not become a uniform architecture-wide failure when the same initialization is reused across targets. Because a median can hide an inconsistent tail, Appendix Table A.3 reports, for every source-target pair, how many of the 45 held-out cases exceed $1.05\times$ and $10\times$, respectively, in the same row/column layout as Table 2. Two structured patterns emerge from that table.

**(1) Cross-architectural comparison.** UNet1 is uniquely and broadly fragile as a transfer *target*: regardless of which model generated the attacked initialization, between 19 and 29 of the 45 cases (42–64%) exceed the notable-transfer threshold when evaluated on UNet1, and 5 to 13 of those cases (11–29%) exceed 10 ×, including when the initialization was produced by an unrelated nn4topopt-style or TopologyGAN-style source. No other target shows this pattern: UNet0 receives at most 8/45 notable transfers from any source, and the NN and GAN targets receive at most 16/45 notable transfers from a source outside their own family, with severe (> 10 ×) transfer almost always confined to a model attacking itself. UNet1 is therefore an architecture-wide weak point in the sense that *any* attacked initialization is disproportionately likely to degrade it, whereas the NN and GAN families are comparatively insulated from initializations crafted against a different architecture.

**(2) Influence of gradient-conditioning depth.** Within the U-Net family, the depth at which the source or target model was conditioned strongly modulates transfer severity: $k = 1$ (UNet1) is the most transfer-fragile depth by a wide margin (as above), $k = 5$ (UNet5) is moderately susceptible but rarely severe (10–22/45 notable cases, almost all below 10 ×), and $k = 0$ (UNet0) is the most insulated (3–8/45 notable cases). The same ordering - $k = 1$ more transfer-fragile than $k = 0$ or $k = 5$ - recurs, in a weaker form, within the NN and GAN families (e.g., NN1 receives 13/45 notable transfers from NN0 and NN5, versus at most 7/45 from any other family, and GAN1 receives 16/45 from NN1 versus at most 10/45 from other sources). This mirrors the non-monotone direct-attack pattern of Table 1: one gradient-conditioning channel is consistently the most exploitable depth across all three architecture families, though the U-Net family is by far the most severely affected.

Fig. 7 shows three transfers into the common UNet1 target on the same physical setup. The UNet5 perturbation visibly weakens the transferred UNet1 prediction, while the NN1 and GAN1 transfers are milder for this case. Rerunning SIMP from the same attacked initializations produces layouts close to the SIMP reference, previewing the physics re-run result of Section 5.4.

**Table 2.** Direct transfer of attacked initializations across surrogate targets. Rows identify the source model and attack used to generate $x_{\text{init}}$; columns identify the target surrogate evaluated with that same initialization. Entries are the median paired ratio of target-model attacked to clean thresholded compliance ($R_{\text{pair}}$, Eq. 16); values above 1.0 × indicate mechanical degradation of the target model under the transferred initialization.

| Source | Attack | UNet0 | UNet1 | UNet5 | NN0 | NN1 | NN5 | GAN0 | GAN1 | GAN5 |
|---|---|---|---|---|---|---|---|---|---|---|
| UNet0 | PGD | 1.48 × | 1.01 × | 1.01 × | 0.98 × | 1.00 × | 1.00 × | 1.01 × | 1.00 × | 1.00 × |
| UNet1 | PGD | 0.98 × | 132.65 × | 1.01 × | 0.99 × | 1.01 × | 1.00 × | 1.00 × | 1.00 × | 1.01 × |

| Source | Attack | UNet0 | UNet1 | UNet5 | NN0 | NN1 | NN5 | GAN0 | GAN1 | GAN5 |
|---|---|---|---|---|---|---|---|---|---|---|
| UNet5 | RAE-PGD | 0.90 × | 1.13 × | 1.38 × | 0.99 × | 1.00 × | 1.01 × | 1.00 × | 1.00 × | 1.00 × |
| NN0 | PGD | 0.98 × | 1.07 × | 1.03 × | 1.03 × | 1.00 × | 1.00 × | 1.00 × | 1.00 × | 1.01 × |
| NN1 | PGD | 0.95 × | 1.16 × | 1.02 × | 0.99 × | 1.16 × | 1.01 × | 1.01 × | 1.02 × | 1.00 × |
| NN5 | Adam | 0.94 × | 1.12 × | 1.04 × | 0.97 × | 1.00 × | 1.11 × | 1.00 × | 1.01 × | 1.01 × |
| GAN0 | Adam | 0.85 × | 1.05 × | 1.04 × | 0.97 × | 1.00 × | 1.00 × | 4.62 × | 1.00 × | 1.00 × |
| GAN1 | PGD | 0.87 × | 1.10 × | 1.02 × | 0.98 × | 1.00 × | 1.00 × | 1.00 × | 36.19 × | 1.00 × |
| GAN5 | PGD | 0.90 × | 1.08 × | 1.05 × | 0.97 × | 1.00 × | 1.00 × | 0.99 × | 1.00 × | 1.07 × |

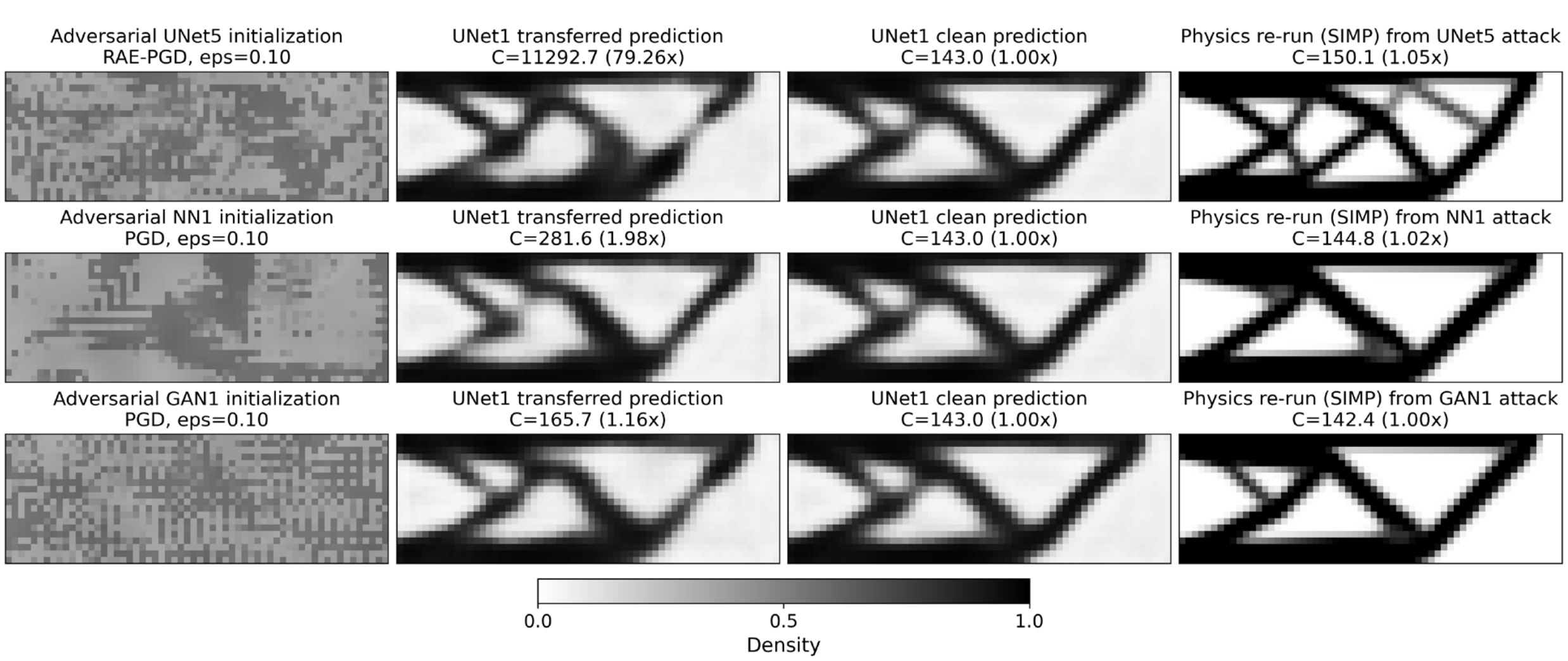

**Fig. 7.** Direct-transfer examples followed by physics-based optimization initialized with attacked density fields. Rows show source attacks transferred to UNet1. Columns show the adversarial

initialization, transferred UNet1 prediction, clean UNet1 prediction, and the SIMP rerun from the same attacked initialization.

*5.4 Adversarial Initializations against Physics-based TO*

The purpose of this section is to understand the robustness of physics-based TO (SIMP) to the same adversarial initializations that degrade the DL surrogates in Sections 5.2–5.3. The surrogate failures documented above do not by themselves imply that the original numerical optimizer fails: in the physics re-run experiment, each attacked initialization is instead handed to SIMP with the same load, supports, volume target, and solver settings, so that only the starting density field differs from the reference (clean) run.

Across the nine reported sources in Table 3, every SIMP-rerun median returns to $1.000\times$ relative to the SIMP reference, matching the clean baseline compliance to three decimal places for every architecture and conditioning depth tested - including UNet1 and GAN1, whose standalone surrogate predictions were degraded by $132.65\times$ and $36.19\times$, respectively (Table 1). The physics solver is therefore highly, though not perfectly, robust to this class of attacked initialization. Pooled across all nine sources and 45 test problems (405 source-instance pairs), 14 cases (3.5%) still exceed a $1.05\times$ compliance penalty after the SIMP rerun, and 6 cases (1.5%) exceed $1.10\times$; the single worst case reaches $1.353\times$ (source UNet0, Table 3). These counts are small but not zero: SIMP can fail to fully wash out an adversarial initialization, and it does so with low but non-negligible probability, most often for the sources with the most exceedances in Table 3 (2/45 for NN5; 1/45 each for UNet0, NN1, GAN1, and GAN5). The correct claim is therefore that SIMP is *more robust* to this non-intrusive channel than the feed-forward surrogates, not that it is *immune* to it.

This robustness is specific to attacks confined to $x_{\text{init}}$. As detailed in Appendix A.5, when the same class of attacker is instead granted access to the solver-internal channels - the compliance-sensitivity field, the sensitivity filter, or the applied load vector - SIMP itself fails to converge to a physically reasonable design, with final compliance degraded by roughly $5\times$ to over $700\times$ depending on which channel is attacked. The boundary between the two results is exactly the boundary between the two threat models defined in Section 3: SIMP's iterative equilibrium solve can re-derive a good design from a bad starting point, but it cannot compensate for corrupted physics inputs that it is required to trust at every iteration.

Taken together, these results support a specific, bounded conclusion: for the non-intrusive, $x_{\text{init}}$-only threat model studied in this paper, keeping the classical SIMP optimizer in the loop - using the learned surrogate as a proposal or warm-start generator rather than as an unchecked final output - very substantially, though not completely, mitigates the demonstrated surrogate vulnerability. It does not extend to a claim that TO software is robust to attacks that reach the solver's internal sensitivity, filtering, or load-vector computations; that channel remains an open exposure documented separately in Appendix A.5.

Table 3 summarizes the aggregate SIMP re-run outcomes, while Fig. 8 illustrates representative cases using the same attacked initializations. Although the attacked surrogate predictions exhibit broken or weakened load paths, the corresponding SIMP re-runs generally recover connected structures close to the reference topology.

**Table 3.** Physics re-run (SIMP) outcomes from directly attacked initializations. The surrogate column gives the median paired adv/clean degradation before the physics rerun. The remaining columns summarize the SIMP rerun from the same attacked initialization over the 45 held-out test problems: the median and maximum ratio of SIMP-rerun to ground-truth (GT) compliance, and the count of the 45 cases exceeding a $1.05\times$ (notable, as in Section 5.3) and a $1.10\times$ (severe) ratio. A ratio of $1.0\times$ means the rerun exactly matches the SIMP reference; values above $1.0\times$ indicate residual degradation that the physics rerun did not fully remove.

| Source | Attack | Surrogate adv/clean | SIMP-rerun/GT (median) | $> 1.05\times$ | $> 1.10\times$ | Maximum |
|---|---|---|---|---|---|---|
| UNet0 | PGD | $1.48\times$ | $1.000\times$ | 1/45 | 1/45 | $1.353\times$ |
| UNet1 | PGD | $132.65\times$ | $1.000\times$ | 1/45 | 0/45 | $1.083\times$ |
| UNet5 | RAE-PGD | $1.38\times$ | $1.000\times$ | 2/45 | 0/45 | $1.067\times$ |
| NN0 | PGD | $1.03\times$ | $1.000\times$ | 1/45 | 0/45 | $1.088\times$ |
| NN1 | PGD | $1.16\times$ | $1.000\times$ | 1/45 | 1/45 | $1.105\times$ |
| NN5 | Adam | $1.11\times$ | $1.000\times$ | 3/45 | 2/45 | $1.304\times$ |
| GAN0 | Adam | $4.62\times$ | $1.000\times$ | 1/45 | 0/45 | $1.084\times$ |
| GAN1 | PGD | $36.19\times$ | $1.000\times$ | 2/45 | 1/45 | $1.118\times$ |
| GAN5 | PGD | $1.07\times$ | $1.000\times$ | 2/45 | 1/45 | $1.229\times$ |

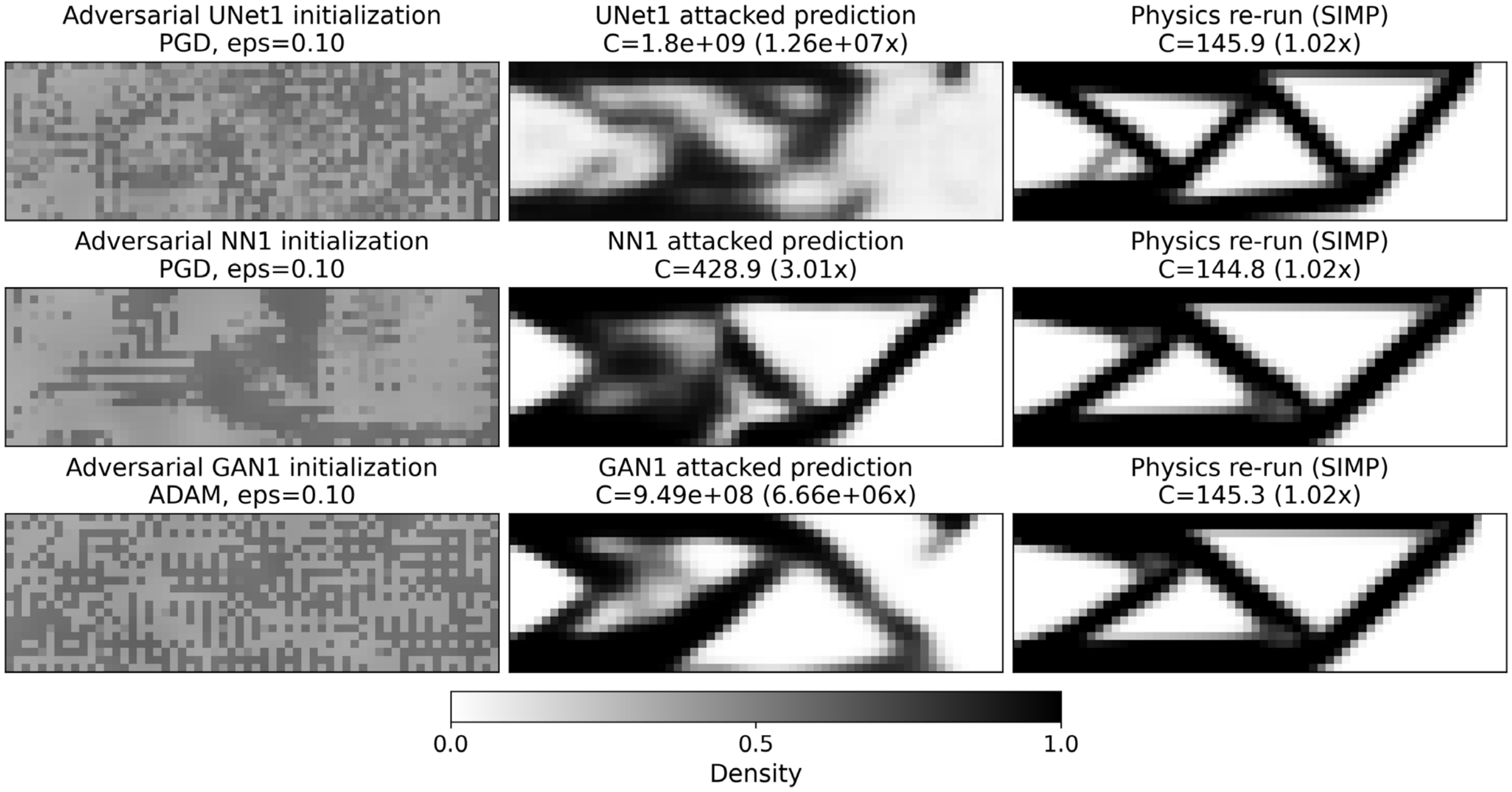

**Fig. 8.** Topological optimization outcomes initialized with attacked density configurations. Rows use the same attacked initializations; columns show the adversarial initialization, the attacked surrogate prediction, and the SIMP rerun.

## 6 Conclusion

This paper presented a mechanics-grounded reliability evaluation of deep learning surrogates acting as autonomous design agents in topology optimization under bounded, non-intrusive perturbations to the initial density channel, $x_{\text{init}}$. Across diverse surrogate architecture families (U-Net, convolutional encoder-decoders, and generative adversarial models), the empirical findings demonstrate that high nominal clean accuracy and near-optimal clean compliance fail to guarantee physical robustness. Bounded initialization perturbations ($\epsilon = 0.10$) can induce severe mechanical compliance degradation, reaching multiple orders of magnitude in vulnerable models (up to $132.65\times$ in the median case, and considerably higher for individual test instances), by triggering localized topological disconnections and severing primary load-bearing members. Furthermore, incorporating richer compliance-gradient physics channels does not confer monotonic robustness across model families, highlighting that architectural inductive biases and training dynamics strongly govern adversarial vulnerability.

A crucial system-level insight emerges from contrasting feed-forward surrogate predictions with physics-in-the-loop optimization. While learned surrogates operate as pure feed-forward approximations without internal equilibrium enforcement, rendering them susceptible to input-level perturbations, re-engaging the classical SIMP numerical solver from these perturbed initializations restores near-optimal structural performance in the large majority of cases (median SIMP-rerun compliance ratio of $1.000\times$), though not universally: a small minority of instances still retain measurable residual degradation after the rerun (Section 5.4). Because SIMP repeatedly resolves finite-element equilibrium and filters compliance sensitivities across iterations, the physics solver tolerates this class of initialization artifact far better than the feed-forward surrogates do, provided external boundary conditions, load vectors, and solver routines remain uncorrupted.

These findings carry vital implications for the design and deployment of learned design agents within automated Industry 4.0 and cyber-physical manufacturing systems. Evaluating data-driven topology optimization models solely through standard computer vision metrics - such as pixel accuracy, MSE, or IoU - is fundamentally insufficient; finite-element compliance verification must be an integral component of model benchmarking. Nor should deep learning surrogates be deployed as unverified, standalone black-box layout generators in mechanically critical environments: they provide maximum utility and physical reliability instead when positioned as rapid proposal engines, generative exploration tools, or warm-start initializers coupled to downstream physics-driven verification and refinement loops.

Looking forward, establishing end-to-end resilience in data-driven structural design opens several promising avenues for future research. Key frontiers include developing physics-informed adversarial training formulations to regularize surrogate sensitivity, investigating higher-resolution and 3D structural benchmarks, and extending reliability evaluations to modern generative paradigms such as diffusion and flow-matching models. In parallel, establishing rigorous anomaly-detection and geometric filtering defenses across digital data-exchange boundaries will be essential to ensure robust, secure, and trustworthy automated design pipelines.

**CRediT authorship contribution statement**

**Hoang Anh Nguyen**: Writing – original draft, Visualization, Validation, Software, Methodology, Investigation, Conceptualization. **Yuan Hong**: Writing - review & editing, Validation, Supervision, Methodology, Conceptualization. **Hongyi Xu**: Writing - review & editing, Validation, Supervision, Methodology, Conceptualization

**Declaration of competing interest**

The authors declare that they have no known competing financial interests or personal relationships that could have appeared to influence the work reported in this paper.

**Acknowledgments**

The authors gratefully acknowledge financial support from the National Science Foundation (CMMI-2434384, CMMI-2142290, CNS-2308730, and CMMI-2326341).

**Appendix**

*A.1 Clean surrogate model detail*

**Table A.1.** Reported clean-test detail.

| Model | Accuracy | IoU | F1 | Median clean/GT |
|---|---|---|---|---|
| UNet0 | 94.7% | 0.889 | 0.939 | 1.04 × |
| UNet1 | 95.0% | 0.895 | 0.942 | 1.00 × |
| UNet5 | 95.3% | 0.902 | 0.946 | 1.01 × |
| NN0 | 93.5% | 0.867 | 0.926 | 1.05 × |
| NN1 | 94.2% | 0.881 | 0.934 | 1.00 × |
| NN5 | 95.5% | 0.905 | 0.948 | 1.00 × |
| GAN0 | 95.2% | 0.900 | 0.944 | 1.00 × |
| GAN1 | 95.3% | 0.902 | 0.946 | 1.00 × |
| GAN5 | 95.3% | 0.903 | 0.946 | 1.00 × |

*A.2 Direct-attack severity distribution*

Table 1 reports only the median paired adv/clean compliance ratio for each model and attack, which cannot by itself show how consistently an attack degrades the held-out test set. Table A.2 reports, for the attack selected per model (the bold entries of Table 1), the count of the 45 held-out test problems whose paired ratio exceeds 2 ×, 10 ×, and 100 ×, respectively.

The vulnerable models fail on a majority of the test set rather than in a few isolated cases: UNet1 and GAN1 both exceed 100 × on more than 40% of held-out problems, and even the least severe vulnerable configuration (UNet0) exceeds 2 × on more than a third of cases. The nn4topopt-style models show the opposite pattern: even their most-affected configuration stays under a quarter of the test set at the 2 × tier, and 10 ×-severity failures are confined to a single case at any tested depth.

**Table A.2.** Direct-attack exceedance counts for the nine reported models. Counts are out of the 45 held-out test problems, using the attack selected per model in Table 1.

| Source | Attack | Median adv/clean | > 2 × | > 10 × | > 100 × | Maximum |
|---|---|---|---|---|---|---|
| UNet0 | PGD | 1.48 × | 17/45 | 11/45 | 9/45 | 1.13e7 × |

| Source | Attack | Median adv/clean | $>2\times$ | $>10\times$ | $>100\times$ | Maximum |
|---|---|---|---|---|---|---|
| UNet1 | PGD | $132.65\times$ | 31/45 | 26/45 | 25/45 | $1.62e7\times$ |
| UNet5 | RAE-PGD | $1.38\times$ | 18/45 | 14/45 | 5/45 | $3.45e6\times$ |
| NN0 | PGD | $1.03\times$ | 6/45 | 1/45 | 0/45 | $10.43\times$ |
| NN1 | PGD | $1.16\times$ | 10/45 | 1/45 | 1/45 | $191.5\times$ |
| NN5 | Adam | $1.11\times$ | 5/45 | 1/45 | 1/45 | $420.1\times$ |
| GAN0 | Adam | $4.62\times$ | 24/45 | 21/45 | 18/45 | $5.53e7\times$ |
| GAN1 | PGD | $36.19\times$ | 35/45 | 27/45 | 20/45 | $2.64e7\times$ |
| GAN5 | PGD | $1.07\times$ | 10/45 | 5/45 | 4/45 | $2.36e6\times$ |

*A.3 Transfer attack success-rate statistics*

Table 2 reports only median transfer ratios. Table A.3 reports, for every source-target pair and using the same attack selected in Table 1, the count of the 45 held-out test problems for which the target's attacked/clean compliance ratio exceeds the notable-transfer threshold of $1.05\times$, and, separately, the count exceeding $10\times$, formatted as "$n_{>1.05\times}/n_{>10\times}$" (each out of 45). Diagonal cells (in bold) are computed from the same underlying per-sample distributions as Table A.2, re-expressed under the $1.05\times/10\times$ thresholds used for the transfer comparison rather than Table A.2's $2\times/10\times/100\times$ tiers, and correspond to Table 2's bold diagonal. (The $>10\times$ counts happen to be identical in both tables, since that threshold is shared.).

**Table A.3.** Transfer exceedance counts across surrogate targets. Entries report $n_{>1.05\times}/n_{>10\times}$ out of the 45 held-out test problems for the target model's attacked/clean compliance ratio.

| Source | Attack | UNet0 | UNet1 | UNet5 | NN0 | NN1 | NN5 | GAN0 | GAN1 | GAN5 |
|---|---|---|---|---|---|---|---|---|---|---|
| UNet0 | PGD | **33/11** | 19/9 | 10/1 | 2/0 | 3/0 | 2/0 | 1/0 | 5/1 | 1/0 |
| UNet1 | PGD | 8/2 | **41/26** | 12/1 | 4/0 | 4/0 | 6/0 | 5/0 | 5/0 | 4/0 |

| Source | Attack | UNet 0 | UNet 1 | UNet 5 | NN 0 | NN 1 | NN 5 | GAN 0 | GAN 1 | GAN 5 |
|---|---|---|---|---|---|---|---|---|---|---|
| UNet 5 | RAE-PGD | 4/1 | 26/11 | **36/14** | 1/0 | 4/0 | 6/0 | 2/0 | 5/2 | 3/1 |
| NN0 | PGD | 7/2 | 26/5 | 19/2 | **19/1** | 13/0 | 8/0 | 5/0 | 5/0 | 8/0 |
| NN1 | PGD | 7/2 | 29/10 | 20/1 | 4/0 | **28/1** | 13/0 | 8/1 | 16/0 | 4/0 |
| NN5 | Adam | 6/0 | 26/13 | 21/3 | 3/0 | 13/0 | **27/1** | 2/1 | 10/3 | 11/1 |
| GAN 0 | Adam | 4/0 | 23/12 | 18/2 | 1/0 | 3/0 | 1/0 | **38/21** | 6/2 | 4/0 |
| GAN 1 | PGD | 3/1 | 24/12 | 14/2 | 2/0 | 4/0 | 4/0 | 6/0 | **44/27** | 2/0 |
| GAN 5 | PGD | 6/0 | 24/13 | 22/2 | 0/0 | 7/0 | 6/0 | 2/1 | 9/1 | **25/5** |

*A.4 Perturbation-budget sensitivity*

The main text uses $\epsilon = 0.10$ so that direct attack, transfer, and SIMP physics re-run are all compared under one common perturbation budget. Table A.4 asks whether the direct-attack story depends on that single number. The attack method is not reselected in this sweep; each row keeps the method used in the main $\epsilon = 0.10$ comparison and changes only the perturbation budget.

The pattern is useful for interpretation. At 0.02 and 0.05, most median ratios remain close to the clean prediction. At 0.10, the vulnerable U-Net and TopologyGAN-style rows separate clearly from the more stable nn4topopt-style rows. At 0.15, some U-Net and TopologyGAN-style cases enter the near-singular thresholded-layout regime, producing very large medians. For this reason, 0.10 is kept as the main budget: it exposes the surrogate failure before relying on the most extreme budget in the sweep.

**Table A.4.** Budget sensitivity for selected direct attacks**.** Cells report median paired adv/clean compliance ratios over the 45 held-out problems.

| Source | Selected attack | $\epsilon = 0.02$ | $\epsilon = 0.05$ | $\epsilon = 0.10$ | $\epsilon = 0.15$ |
|---|---|---|---|---|---|
| UNet0 | PGD | $0.99\times$ | $1.00\times$ | $1.48\times$ | $1.76\times$ |

| Source | Selected attack | $\epsilon = 0.02$ | $\epsilon = 0.05$ | $\epsilon = 0.10$ | $\epsilon = 0.15$ |
|---|---|---|---|---|---|
| UNet1 | PGD | $1.03\times$ | $1.18\times$ | $132.65\times$ | $4.19e6\times$ |
| UNet5 | RAE-PGD | $1.00\times$ | $1.05\times$ | $1.38\times$ | $3.13\times$ |
| NN0 | PGD | $1.01\times$ | $1.02\times$ | $1.03\times$ | $0.96\times$ |
| NN1 | PGD | $1.01\times$ | $1.05\times$ | $1.16\times$ | $1.31\times$ |
| NN5 | Adam | $1.00\times$ | $1.01\times$ | $1.11\times$ | $1.19\times$ |
| GAN0 | Adam | $1.01\times$ | $1.07\times$ | $4.62\times$ | $240.13\times$ |
| GAN1 | PGD | $1.01\times$ | $1.70\times$ | $36.19\times$ | $3.92e6\times$ |
| GAN5 | PGD | $1.00\times$ | $1.00\times$ | $1.07\times$ | $1.27\times$ |

*A.5 Intrusive solver-channel attacks*

The physics re-run result in Section 5.4 has a narrow meaning. It says that SIMP can repair the attacked initial-density fields when the physical problem and optimizer are left unchanged. It does not say that SIMP can repair an attack that changes the physical inputs or the update calculation itself. That non-intrusive case is already established in depth by Section 5.4 and Table 3, across 45 held-out test problems and nine attack sources: the classical optimizer washes out an adversarial $x_{\text{init}}$ almost completely, with the SIMP-rerun median returning to $1.000\times$ the reference compliance for every source tested. Table A.5 below instead probes the opposite case: what happens once the attacker is allowed inside the solver loop or the physical problem definition, rather than the initial guess alone.

A phantom-load attack adds a persistent random force perturbation to 5% of the free degrees of freedom, plus independent Gaussian jitter re-sampled every iteration (both at $\epsilon = 0.5$, five times the $\epsilon = 0.10$ bound used for the non-intrusive channel throughout this paper), directly corrupting the load vector that drives the equilibrium solve. Sensitivity poisoning adds fresh zero-mean Gaussian noise, with standard deviation $0.5\times$ the maximum sensitivity magnitude, to the raw compliance-sensitivity field at every iteration, corrupting the information used by the optimality-criteria update. Filter poisoning adds a deterministic checkerboard pattern, sized at $0.5\times$ the maximum filtered-sensitivity magnitude, directly to the filter's output; because this pattern is exactly what the sensitivity filter exists to suppress, it is a worst-case rather than generic corruption of that regularization step. Changing support or boundary-condition data would belong to the same intrusive class because it would redefine the boundary-value problem rather than merely changing an initialization. Unlike the non-intrusive channel's single bounded perturbation, each of these three corruptions is re-applied at every solver iteration, which is why they can prevent convergence outright rather than merely biasing the result. These attacks are not part of the learned-surrogate benchmark; they are included only to substantiate the claim in Sections 3 and 5.4 that the intrusive channel, once available to an attacker, can defeat the physics solver itself.

**Table A.5.** Scope check for intrusive Top88/SIMP attacks. All runs use the same $20 \times 60$ half-MBB setup with target volume 0.5.

| Configuration | Channel changed | Final compliance | Iterations | Outcome |
|---|---|---|---|---|
| Baseline | none | ~203.3 | 94 | converged |
| Sensitivity poisoning | compliance sensitivities | ~1127.6 | 100 | failed to converge |
| Phantom load injection | load vector | ~56,949.0 | 100 | failed to converge |
| Filter poisoning | sensitivity filter output | ~147,017.8 | 6 | diverged/locked |

The lesson is the separation between two failure modes. Feed-forward surrogates can be sensitive to a bounded $x_{\text{init}}$ perturbation because they do not re-solve mechanics during prediction. SIMP, by contrast, can wash out that initialization perturbation when the physics and update rule are intact (Section 5.4, Table 3). But if the attacker changes loads, supports, sensitivities, filters, or other solver-facing quantities, the optimizer is no longer solving the same trusted problem: the three channels above push final compliance from roughly $5 \times$ to over $700 \times$ the clean baseline. The physics re-run result in the main text should therefore be read only as an $x_{\text{init}}$-channel result, not as a general robustness claim for corrupted physical inputs or solver internals.

**Data Availability**

Upon acceptance, the data and model for reproducing the results will be made available upon email request.

**Acknowledgement**

The authors gratefully acknowledge financial support from the National Science Foundation (CMMI-2434384, CMMI-2142290, CNS-2308730, and CMMI-2326341).